\documentclass[12pt,a4paper]{article}
\usepackage{graphicx}
\usepackage{fullpage}
\usepackage[ruled,vlined]{algorithm2e}
\usepackage{setspace}
\usepackage{natbib}

\title{Rough Sets Computations to Impute Missing Data}
\author{Fulufhelo Vincent Nelwamondo and Tshilidzi Marwala\\
\small School of Electrical and Information Engineering,\\ \small University of the Witwatersrand, \\ \small Private Bag 3, \small Wits,   \small 2050, \small South Africa}

\begin{document}
\doublespacing
\maketitle

\begin{abstract}

Many techniques for handling missing data have been proposed in the literature. Most of these techniques are overly complex. This paper explores an imputation technique based on rough set computations. In this paper, characteristic relations are introduced to describe incompletely specified decision tables.It is shown that the basic rough
set idea of lower and upper approximations for incompletely
specified decision tables may be defined in a variety of different
ways.  Empirical results obtained using real data are given and they provide a valuable and promising insight to the problem of missing data. Missing data were predicted with an accuracy of up to 99\%.
\end{abstract}

\section{Introduction}

There are three general ways that have been used to deal with the problem of missing data \citep{book:Little}. The simplest method
is known as `listwise deletion' which, simply deletes
instances with missing values. The major disadvantage of this
method is the dramatic loss of information in data sets. \cite{journal:Kim_Curry} found that when 2\% of the features are
missing and the complete observation is deleted, up to 18 percent
of the total data may be lost. The second common technique imputes the data by finding estimate
of the values and missing entries are replaced with these
estimates.  Various estimates have been used and these estimates include zeros, means
and other statistical calculations. These estimations are then
used as if they were the observed values. Another common technique assumes some models for the prediction of the missing values and uses the maximum likelihood
approach to estimate the missing values.

A graet deal of research has been conducted to find new ways of
approximating the missing values. Among others, 
\cite{journal:Abdella} and 
\cite{Shakir} have used neural networks together with Genetic
Algorithms (GA) to approximate missing data. 
\cite{Qiao} have used neural networks and Particle Swarm
Optimization (PSO) to keep track of the dynamics of a power plant
in the presence of missing data.  
\cite{Nauck_and_Kruse} and \cite{Journal:Gabrys} have  used
Neuro fuzzy for learning in the presence of missing data. A
different approach was taken by \cite{journal:Wang2005} who
replaced incomplete patterns with fuzzy patterns. The patterns
without missing values are, along with fuzzy patterns, used to
train the neural network. In his model, the neural network learns
to classify without actually predicting the missing data. Special
attention in the literature has been given to imputation techniques
such as the Expectation maximisation as well as the use of neural
networks, coupled with an optimisation technique such as genetic
algorithms. The use of neural networks  comes with a greater cost
in terms of computation and in that data has to be made available
before the missing condition occurs.  This paper proposes a new algorithm based on rough set theory for missing data estimation. Although other simmillar methods have been mentioned in the literature \citep{Nakata,GrzymalaBusse_Conf}, this paper also applies a  rough set technique for missing data imputation to a large and real database for the first time.  It is envisaged in this work  that in large databases, it is more likely that the
missing values could be correlated to some other variables
observed somewhere in the same data. Instead of approximating
missing data, it might therefore be cheaper to spot similarities
between the observed data instances and those that contain missing
attributes.

\section{Applications of Rough Sets}

There are many applications of rough sets reported in literature.
Most of  the applications assume that complete data is available
\citep{GrzymalaBusse_Conf}. This is, however, not often the case in
real life situations. There is also a great deal of information
regarding various applications of rough sets in medical data sets.
Rough sets have been used mostly in prediction cases and  \cite{journal:Rowland} compared neural networks and
rough sets for the prediction of ambulation following a spinal
cord injury. Although rough sets performed slightly lower than
neural networks, they proved that they can still be used in
prediction problems. Rough sets have also been used in  learning
Malicious Code Detection \citep{WCCI_Zhang} and in Fault diagnosis
\citep{TayShen}.  \cite{Grzymala-Busse:miss} have
presented nine approaches of imputing up missing values. Among
others, the presented methods include selecting the most common
attribute, \emph{concept most common} attribute, assigning all
possible values related to the current concept, deleting cases
with missing values, treating missing values as special values and
imputing for missing values using other techniques such as neural
networks, and maximum likelihoods approaches. Some of the
techniques proposed come with expense either in terms of
computation time or loss of information.

\section{Rough Set Theory}

The rough sets theory provides a technique of reasoning from vague
and imprecise data \citep{GohLaw}. The technique is based on the
assumption that some information is associated somehow with
\emph{some information} of the universe of the discourse
\citep{book:Komorowski_et_al,Yang_John}. Objects with the
same information are \emph{indiscernible} in the view of the
available information. An elementary set consisting of
indiscernible objects forms a basic granule of knowledge. A union
of elementary set is referred to as a crisp set, otherwise the set
is considered to be  rough. The next few subsections briefly
introduce concepts that are common to rough set theory.

\subsection{Information System}

An information system ($\Lambda$), is defined as a pair
$(\textbf{U},A)$ where $\textbf{U}$ is a finite set of objects
called the universe and $A$ is a non-empty finite set of
attributes  as shown in Eq \ref{IS} below \citep{Yang_John}.

\begin{equation}\label{IS}
    \Lambda=(\textbf{U},A)
\end{equation}

Every attribute $a \in A$ has a value which  must be a member of a
value set $V_a$ of the attribute $a$.

\begin{equation}\label{VS}
    a:\textbf{U} \to V_a
\end{equation}

A rough set is defined  with a set of attributes and the
indiscernibility relation between them. Indiscernibility is
discussed next.

\subsection{Indiscernibility Relation}

Indiscernibility relation is one of the fundamental ideas of rough
set theory \citep{Grzymala-Busse}. Indiscernibility simply implies
similarity \citep{GohLaw}.  Given an information system $\Lambda$
and subset $B \subseteq A$, $B$ determines a binary relation
$I(B)$ on $\textbf{U}$:

\begin{equation}\label{IR}
    (x,y) \in I(B) \quad iff \quad a(x)=a(y)
\end{equation} for all $a \in B$ where $a(x)$ denotes the value of attribute $a$
for element $x$. Eq (\ref{IR}) implies that any two elements that
belong to $I(B)$ should be identical from the point of view of
$a$. Suppose $\textbf{U}$ has a finite set of $N$ objects
$\{x_1,x_2,\dots,x_N\}$. Let $Q$ be a finite set of $n$ attributes
$\{q_1,q_2,\dots,q_n\}$ in the same information system $\Lambda$,
then,

\begin{equation}\label{IR2}
    \Lambda=\langle\textbf{U},Q,V,f \rangle
\end{equation}
 where $f$ is the \emph{total decision function} called the information
 function. From the definition of Indiscernibility Relation given in  this section, any two objects have a  similarity relation to attribute $a$ if they have the same
 attribute values everywhere except for the missing values.

\subsection{Information Table and Data Representation}

An Information Table (IT) is used in rough sets theory as a way of
representing the data. The data in the IT are arranged based on
their condition attributes and decision attribute ($\mathcal{D}$).
Condition attributes and decision attribute are analogous to the
independent variables and dependent variable \citep{GohLaw}. These
attributes are divided into $C \cup \mathcal{D}=Q$ and $C \cap
\mathcal{D}=\emptyset$. An IT can be classified into complete and
incomplete classes. All objects in a complete class have known
attribute values whereas an IT is considered incomplete if at
least one attribute variable has a missing value. An example of an
incomplete IT is given in Table \ref{SampleTAB}.

\begin{table}[!hbt]
\begin{center}
\setlength{\tabcolsep}{10pt} \caption{An example of an Information
Table with missing values}\label{SampleTAB}
\begin{tabular}{|c|ccc|c|}
\hline
 {} & $x_1$ & $x_2$ &$x_3$ &$\mathcal{D}$\\
 \hline \hline

1&1& 1  & 0.2  &  B  \\
 2 &1 & 2 & 0.3& A \\
3 &0& 1 & 0.3  &  B \\
4  &?& ?& 0.3& A  \\
5  &0& 3  & 0.4& A \\
6  &0& 2  & 0.2  & B \\
7 &1&  4  & ?  &  A\\

\hline
\end{tabular}
\end{center}
\end{table}

Data is represented by a table where each row represents an
instance, sometimes referred to as an object. Every column
represents an attribute which can be a measured variable. This
kind of a table is also referred to as Information System
\citep{book:Komorowski_et_al}.

\subsection{Decision Rules Induction}

Rough sets also involve generating decision rules for a given IT.
The rules are normally determined based on condition attributes
values \citep{GohLaw}. The rules are presented in an \emph{if}
CONDITION(S)-\emph{then} DECISION format. This paper will not
directly focus on rule induction since the major interest of this
work is to estimate the missing data as opposed to taking the
decision.

\subsection{Set Approximation}

There are various properties of rough sets that have been
presented in \citep{book:Pawlak} and \citep{journal:Pawlak.}. Some
of the properties are discussed below.

\subsubsection{Lower and Upper Approximation of Sets}
The lower and upper approximations are defined on the basis of
indiscernibility relation discussed above. The lower approximation
is defined as the collection of cases whose equivalent classes are
contained in the cases that need to be approximated whereas the
upper approximation is defined as the collection of classes that
are partially contained in the set that needs to be approximated
\citep{journal:Rowland}.

Let \textbf{concept} $X$ be defined as a set of all cases defined
by a specific value of the decision. Any finite union of
elementary set, associated with $B$ is called a $B-definable$ set
\citep{Grzymala-Busse}. The set $X$ is approximated by two
$B-definable$ sets, referred to as the B-lower approximation
denoted by $\underline{B}X$  and B-upper approximation,
$\overline{B}X$. The B-lower approximation is defined as
\citep{Grzymala-Busse}

\begin{equation}\label{B-low}
    \{x\in \textbf{U}|[x]_B \subseteq X\}
\end{equation}

and the B-upper approximation is defined as

\begin{equation}\label{B-upp}
    \{x\in \textbf{U}|[x]_B \cap X \neq \emptyset\}
\end{equation}

There are other methods that have been reported in the literature
for defining the lower and upper approximations for a completely
specified decision tables. Some of the common ones include
approximating the lower and upper approximation of $X$ using
Equations \ref{B-low2L} and \ref{B-upp2U} respectively as follows
\citep{GrzymalaBusse_Conf}:

\begin{equation}\label{B-low2L}
    \cup\{[x]_B |x\in \textbf{U},[x]_B \subseteq X\}
\end{equation}

\begin{equation}\label{B-upp2U}
    \cup\{[x]_B |x\in \textbf{U},[x]_B \cap X \neq \emptyset\}
\end{equation}

The definition of definability is modified in cases of
incompletely specified tables. In this case, any finite union of
characteristics sets of $B$ is called a $B-definable$ set. Three
different definitions of approximations have been discussed
\cite{Grzymala-Busse}. Again letting $B$ be a subset of $A$ of all
attributes and $R(B)$ be the characteristic relation of the
incomplete decision table with characteristic sets $K(x)$, where
$x\in U$, the following are defined:

\begin{equation}\label{B-low_inc}
    \underline{B}X=\{x\in \textbf{U} | K_B(x) \subseteq X\}
\end{equation} and

\begin{equation}\label{B-upp_inc}
    \overline{B}X=\{x\in \textbf{U} | K_B(x) \cap X \neq \emptyset\}
\end{equation}

Equations  \ref{B-low_inc} and \ref{B-upp_inc} are referred
to as $singletons$. The $subset$ lower and upper approximations of
incompletely specified data sets are then defined as:

\begin{equation}\label{B-low_inc2}
    \cup\{K_B(x) |x\in \textbf{U},K_B(x) \subseteq X\}
\end{equation} and

\begin{equation}\label{B-upp_inc2}
    \cup\{K_B(x) |x\in \textbf{U},k_B(x) \cap X \neq \emptyset\}
\end{equation}

More information on these methods can be found in
\citep{GrzymalaBusse_Conf,Grzymala-Busse:miss,Grzymala-Busse:LERS,Grzymala-Busse}.

It follows from the properties that a crisp set is only defined if
$\underline{B}(X)=\overline{B}(X)$. Roughness therefore is defined
as the difference between the upper and the lower approximation.

\subsubsection{Rough Membership Functions}

Rough membership function is a function $\mu^x_A:\textbf{U} \to
[0,1]$ that when applied to object $x$, quantifies the degree of
overlap between set $X$ and the indiscinibility set to which $x$
belongs. The rough membership function is used to calculate the
plausibility, defined as

\begin{equation}\label{B-upp_inc3}
    \mu_A^X(X)=\frac{|[x]_B \cap X|}{|[x]_B|}
\end{equation}

\section{Missing Data Imputation Based on Rough Sets}

The algorithm implemented here imputes the missing values by
presenting a list of all possible values, based on the observed
data. As mentioned earlier, the hypothesis here is that in most
finite databases, a case similar to the missing data case could
have been observed before. It therefore should be cheaper to use
such values, instead of computing missing values with complex
methods such as neural networks. The algorithm implemented is
shown in Algorithm \ref{RSDA_alg}, followed by a
\emph{work-through example} demonstrating how the missing values
are imputed. There are two approaches to reconstructing the
missing values. The missing values can either be probabilistically
interpreted or be possibilistically interpreted \citep{Nakata}.

\begin{algorithm}
 \caption{Rough sets based missing data imputation algorithm\label{RSDA_alg}}
 \dontprintsemicolon

 \SetKwData{Left}{left}
 \SetKwData{This}{this}
 \SetKwData{Up}{up}
 \SetKwFunction{Union}{Union}
 \SetKwFunction{FindCompress}{FindCompress}
 \SetKwInOut{Input}{input}
 \SetKwInOut{Output}{output}
 \SetKwInOut{Assumption}{Assumption}

\Input{Incompete data set $\Lambda$ with $a$ attributes and $i$
instances.\\
All these instances should belong to a desision $\mathcal{D}$}

\Output{A vector containing possible missing values}

\Assumption{$\mathcal{D}$ and \emph{some} attributes will always
be known}

\SetLine
 \ForAll i
 {
$\rightarrow$ Partition the input space according to $\mathcal{D}$
$\rightarrow$ Arrange all attributes according to order of availability, with $\mathcal{D}$ being first.\\
 }
\ForEach{$attribute$}
 {
 $\rightarrow$ Without directly extracting the rules, use the available information to extract
relationships to other instances $i$ in the $\Lambda$.\\
$\rightarrow$ The family of equivalent classes $\varepsilon (a)$
containing each object $o_i$ for all input attributes is computed.

$\rightarrow$ The degree of belongingness $\kappa(o[A]
1/|dom(a_{i_{missing}})|$ where $o \neq o'$ and $dom(x_{1_4})$
denotes the domain of attribute $x_{1_4}$, which is the forth
instance of $x_1$, and $|dom(x_{1_4})|$ is the cardinality of
$dom(x_{1_4})$
\While{extracting relationships}{If  $i$ has the same attribute values with $a_j$ everywhere except for the missing value, replace the missing value, $a_{missing}$, with the value $v_j$, from $a_j$, where $j$ is an index to onother instance.\\
Otherwise proceed to the next step\\}

 $\rightarrow$ Complete the lower approximation of each attribute,given the available data of the same instance with the missing value.\\
 \While {doing this}{
 IF more than one $v_j$ values are suitable for the estimation, postpone the replacement for later when it will be clear which value is appropriate\\
    }
$\rightarrow$ Compute the incomplete upper approximations of
    each subset partition.\\

    $\rightarrow$ Do the computation and imputation of missing data as was done with the lower approximation.\\
    $\rightarrow$ Either $crips$ sets will be found, otherwise, $rough$ sets can be used and missing data can be heuristically be selected from the obtained $rough$ set.
 }
\end{algorithm}

In our example, the degree of belongingness
$\kappa(o[x_{1_4}]=o[x_{1_4}]=1/|dom(x_{1_4})|$ where $ o \neq o'$
and $dom(x_{1_4})$ denotes the domain of attribute $x_{1_4}$,which is the forth instance of $x_1$, and $|dom(x_{1_4})|$ is the
cardinality of $dom(x_{1_4})$. If the missing values were to be
possibilistically interpreted, all attributes have the same
possibilistic degree of being the actual one.


The algorithm in this study is fully dependent on the available data and makes
no additional assumptions about the data or the distribution
thereof.  As presented in the algorithm, a list of possible values
is given in a case where a crisp set could not be found. It is
from this list that possible values may be heuristically chosen.
A justification to this is that it is not always the case that we
need to know the  \emph{exact} value. As a result, it may be
cheaper to have a $rough$ value. The possible imputable values are
obtained by collecting all the entries that lead to a particular
decision $\mathcal{D}$. The algorithms used in this application is
a simplified version of the algorithm of 
\cite{Hong_et_al}.

The algorithm  will now be illustrated using an example. Missing
values will be denoted by the question mark  $(?)$ symbol.
Attribute values of attribute $a$ are denoted as $V_a$. Using the
notation defined in \cite{journal:Gediga}, we let $rel_Q(x)$
represent a set of all \emph{Q-relevant attributes} of $x$.
Assuming an IT as presented in Table \ref{SampleTAB}, where
 $x_1$ is in binary form, $x_2 \in [1:5]$ and being integers
and $x_3$ can either be 0.2, 0.3 or 0.4.

The algorithms firstly seeks relationship between variables. Since
this is a small database, it is assumed that the only variable
that will always be known is the decision. The first step will be
to partition the data according to the decision and this could be
done as follows:

\begin{eqnarray*}
\varepsilon (D)=\{o_1,o_3,o_6\},\{o_2, o_4, o_5, o_7\}
\end{eqnarray*}

 Two partitions are obtained due the binary nature of the decision in the chosen example. The next step is to extract indiscernible relationships within each attribute. For $x_1$, the following is obtained:
\begin{eqnarray*}
IND(x_1)=\{(o_1,o_1),(o_1,o_2),(o_1,o_4),(o_1,o_7),(o_2,o_2),(o_2,o_4),(o_2,o_7),\\(o_3,o_3),(o_3,o_4),(o_3,o_5),(o_3,o_6),(o_4,o_4),(o_4,o_5),(o_4,o_6)(o_4,o_7),\\(o_5,o_5),(o_5,o_6),(o_6,o_6),(o_7,o_7)\}\\
\end{eqnarray*}

The family of equivalent classes $\varepsilon (x_1)$ containing
each object $o_i$ for all input variables is computed as follows:
\begin{eqnarray*}
\varepsilon (x_1)=\{o_1,o_2,o_4, o_7\},\{o_3,o_4 o_5, o_6\}
\end{eqnarray*}

Similarly,
\begin{eqnarray*}
\varepsilon (x_2)=\{o_1,o_3,o_4\},\{o_2,o_4, o_6\}
,\{o_4,o_5\},\{o_,o_7\},\{o_4\}\{0_7\}
\end{eqnarray*}
and
\begin{eqnarray*}
\varepsilon (x_3)=\{o_1,o_6, o_7\},\{o_2,o_3, o_4, o_7\},\{o_5,
o_7\}
\end{eqnarray*}

In our example, the degree of belongingness
$\kappa(o[x_{1_4}]=o[x_{1_4}]=1/|dom(x_{1_4})|$ where $o \neq o'$
and $dom(x_{1_4})$ denotes the domain of attribute $x_{1_4}$,
which is the fourth instance of $x_1$, and $|dom(x_{1_4})|$ is the
cardinality of $dom(x_{1_4})$. If the missing values were to be
possibilistically interpreted, each attribute has the same
possibilistic degree of  being the actual one. The lower
approximations is defined as:

\begin{equation}\label{B-low2}
    \underline{A}(X_{miss},\{X_{avail},\mathcal{D}\})=\{E(X_{miss})|\exists(X_{avail},\mathcal{D}), E(X)\subseteq
    (X_{avail},\mathcal{D})\}
\end{equation} whereas the upper approximation is defined as

\begin{equation}\label{B-upp2}
    \overline{A}(X_{miss},\{X_{avail},\mathcal{D}\})=\{E(X_{miss})|\exists(X_{avail},\mathcal{D}),
    E(X)\cap X_{avail}\cap \mathcal{D}\}
\end{equation}

Using $IND(x_1)$, the families of all possible classes containing
$o_4$ are given by

\begin{eqnarray*}
Poss \varepsilon (x_1)_{o_i}=\{o_1,o_2,o_7\},\{o_1,o_2,o_4,o_7\}, i=1,2,7\\
Poss \varepsilon (x_1)_{o_i}=\{o_3,o_5,o_6\},\{o_3,o_4,o_5,o_6\}, i=3,5,6\\
Poss \varepsilon (x_1)_{o_4}=\{o_4},{o_1,o_2,o_7\},\{o_3,o_4,o_5,o_6\}\\
\end{eqnarray*}

The probabilistic degree to which we can be sure that the chosen
value is the right one is given by \citep{Nakata}

\begin{eqnarray*}
\kappa((\{o_i\})\in \varepsilon(x_1))=1/2, i=1,2,7\\
\kappa((\{o_i\})\in \varepsilon(x_1))=1/2, i=3,5,6\\
\kappa((\{o_i\})\in \varepsilon(x_1))=1/2, i=4\\
else\\
\kappa(\{o_i\})\in \varepsilon(x_1))=0
\end{eqnarray*}

The else part applies to all other conditions such as
$\kappa(\{o_1,o_2,o_3\})\in \varepsilon(x_1))=0$. A family of
weighted equivalent classes is now computed as follows:
\begin{eqnarray*}
\varepsilon (x_1)=\{\{o_1,o_2,o_4, o_7\}\{1/2\}\},\{\{o_3,o_4 o_5,
o_6\}\{1/2\}\}
\end{eqnarray*}

The values $\varepsilon (x_2)$ and $\varepsilon (x_3)$ are
computed in a similar way. We then use these families of weighted
equivalent classes to obtain the lower and upper approximations as
presented above. The degree to which the object $o$ has the same
value as object $o'$ on the attributes is referred to as the
degree of belongingness and is defined in terms of the binary
relation for indiscernibility as \citep{Nakata}:

\begin{eqnarray*}
IND(X)=\{((o,o'),\kappa(o[X]=o'[X]))|
(\kappa(o[X]=o'[X])\\
\neq 0)\wedge(o \neq o')\}\cup\{((o,o),1)\}
\end{eqnarray*} where $\kappa(o[X]=o'[X])$ is the indiscernibility degree of the
objects $o$ and $o'$ and this is equal to the degree of
belongingness,

\begin{eqnarray*}
\kappa(o[X]=o'[X])= \stackrel{\otimes}{_{A_i \in X}}
\kappa(o[A_i]=o'[A_i])
\end{eqnarray*} where the operator $\otimes$ depends on  whether the missing
values are possibilistically or probabilistically interpreted. For
probabilistic interpretation, the parameter is a product denoted
by $\times$, otherwise the operator $min$ is used.

%
%

\section{Experimentatal Evaluation}

\subsection{Database}

The data used in this test was obtained from the South African
antenatal sero-prevalence survey of 2001. The data for this survey
is obtained from  questionnaires answered by pregnant women
visiting selected public clinics in South Africa. Only women
participating for the first time in the survey were eligible to
answer the questionnaire.

Data attributes used in this study are the \emph{HIV status,
education level, gravidity, parity, age, age of the father, race}
and \emph{region }. The HIV status is the decision and is
represented in a binary form, where 0 and 1 represent  negative
and positive respectively. Race is measured on the scale 1 to 4
where 1, 2, 3, and 4 represent African, Coloured, White and Asian,
respectively. The data used was obtained in three regions and are
referred to as region A, B and C in this investigation. The
education level was measured using integers representing the
highest grade successfully completed, with 13 representing
tertiary education. Gravidity is the number of pregnancies,
complete or incomplete, experienced by a female, and this variable
is represented by an integer between 0 and 11. Parity is the
number of times the individual has given birth and multiple births
are counted as one. Both parity and gravidity are important, as
they show the reproductive activity as well as the reproductive
health state of the women. Age gap is a measure of the age
difference between the pregnant woman and the prospective father
of the child. A sample of this data set is shown in Table
\ref{SampleHIV}.

\begin{table}[!hbt]
\begin{center}
\caption{Extract of the HIV database used, with missing
values}\label{SampleHIV}
\begin{tabular}{|ccccccc|c|}
\hline\noalign{\smallskip}
  Race & Region &Educ & Gravid &Parity &Age &Father's age   & HIV \\
 \hline \hline

  1& C& ?  & 1  & 2 & 35 & 41&  0\\
  2& B & 13 & 1  & 0  & 20 & 22   & 0 \\
  3& ?& 10 & 2  & 0  & ? &  27   & 1 \\
  2 & C& 12 & 1  & ?  & 20 & 33  & 1 \\
   3&B&9  & ?  & 2  & 25 & 28   & 0\\
   ?&C&9  & 2  & 1  & 26 & 27  & 0\\
   2&A&7  & 1  & 0   &15 & ? & 0 \\
   1&C&? & 4  & ?  & 25& 28   & 0\\
   4& A &7 &  1  & 0 &  15 & 29  & 1 \\
  1& B&11  & 1  & 0 &  20 & 22  & 1\\
\hline
\end{tabular}
\end{center}
\end{table}

\subsection{Data Preprocessing}

As mentioned in a previous section, the HIV/AIDS data that is used in this work is obtained from a survey performed on pregnant women. Like all data in raw form, there are several steps that need to be taken in order to ensure the data is in usable form. There are several types of outliers that have been identified in the data. Firstly, some of the data records were not complete. This is probably due to the fact that the people being surveyed omitted certain information and also errors made by the person who manually recorded the surveys onto a spreadsheet. The outliers were from incorrectly entered variables. For instance \emph{Gravidity} is defined as the number of times a woman has been pregnant and \emph{parity} is described as the number of times a woman has given birth. Any instance whereby the value of parity is greater than that of parity, the whole observation was considered an outlier and was removed. The justification to this is that it is not possible for a woman to give birth more than she has been pregnant.

\subsection{Variable Discretisation}

The discretisation defines the granularity with which we would like to analyse the universe of discourse. If one chooses to discretise the variables into a large number of categories the rules extracted are more complex to analyse. Therefore, if one would like to use rough sets for rule analysis and interpretation rather than for classification it is advisable that the number of categories be as small as possible. For the purposes of this work the input variables have been discretised into four categories. A description of the categories and their definition is shown in Table \ref{categories}.  Table
\ref{SimpleHIV3} shows the simplified version of the information system  shown in Table  \ref{SampleHIV}.

\begin{table}[!htbp]
	\caption{A table showing the discretised variables.}\label{categories}	
	\centering
	\begin{tabular}{|c |c |c |c |c |c |c|}
	\hline
	\textbf{Race} & \textbf{Age} & \textbf{Education} & \textbf{Gravidity} & \textbf{Parity} & \textbf{Father's Age} & \textbf{HIV} \\ \hline \hline
	1 & $\leq 19$ & Zero ($0$) & Low ($\leq 3$) & Low ($\leq 3$) &$\leq 19$ & 0 \\
	2 &  $[20-29$]) & P ($1-7$) & High ($>3$) & High ($>3$)& ($[20-29]$) & 1 \\ 
	3 &  $[30-39$]) & S ($8-12$) & - & - & ($[30-39]$) &  -\\ 
	4 &  $\geq 40$ & T ($13$) &  & - & - $\geq 40 $& - \\  \hline
	\end{tabular}	
	
\end{table}

\begin{table}[!htb]
\begin{center}
\caption{Extract of the HIV database used, with missing
values after discretisation}\label{SimpleHIV3}
\begin{tabular}{|ccccccc|c|}
\hline\noalign{\smallskip}
  Race & Region &Educ & Gravid &Parity &Age &Father's age  & HIV \\
 \hline \hline

  1& C& ?  & $\leq 3$  & $\leq 3$ & [31:40]& [41:50] &  0\\
  2& B & T & $\leq 3$  & $\leq 3$  & $\leq 20$ & [21:30]   & 0 \\
  3& ?& S & $\leq 3$  & $\leq 3$  & ? &  [21:30]    & 1 \\
  2 & C& S & $\leq 3$  & ?  & $\leq 20$ & [31:40]  & 1 \\
   3&B&S  & ?  &$\leq 3$  & [21:30]& [21:30]  & 0\\
   ?&C&S  & $\leq 3$  & $\leq 3$  & [21:30] & [21:30]   & 0\\
   2&A&P & $\leq 3$  & $\leq 3$   &$\leq 20$& ? & 0 \\
   1&C&? & $>3$  & ?  & [21:30]& [21:30]  & 0\\
   4& A &P &  $\leq 3$  & $\leq 3$ &  $\leq 20$ & [21:30]  & 1 \\
  1& B& S  & $\leq 3$  & $\leq 3$ &  $\leq 20$ & [21:30]   & 1\\
\hline
\end{tabular}
\end{center}
\end{table}

\subsection{Results and Discussion}

The experimentation was performed using both the original and the
simplified data sets. Results obtained in both cases are
summarised in Table \ref{RSResultsTAB}.

\begin{table}[!h]
\begin{center}
\setlength{\tabcolsep}{10pt} \caption{Missing data estimation
results for both the original data and the generalised
data}\label{RSResultsTAB}
\begin{tabular}{|c|cccc|}
\hline
 {} & Education & Gravidity & Parity & Father's age\\
 \hline \hline
Original&83.1& 86.5  & 87.8  &  74.7  \\
Generalised &99.3 & 99.2 & 99& 98.5 \\
\hline
\end{tabular}
\end{center}
\end{table}

It can be seen that the prediction accuracy is much higher for the
generalised data set. This is because the states have been
reduced. Furthermore, instead of being exact, the likelihood of
being correct is even higher if one has to give a rough estimate.
For instance, instead of saying that someone has a highest level
of education of 10, it is much safer to say, \emph{They have
secondary education}. Although this approach leaves details, it is
often the case that the left-out details are not required. In a
decision system such as the one considered in this chapter,
knowing that the prospective father is 19 years old may carry the
same weight as saying that the father is a $teenager$.

\section{Conclusion}

Rough sets have been used for missing data imputation and
characteristic relations are introduced to describe incompletely
specified decision tables. It has been shown that the basic rough
set idea of lower and upper approximations for incompletely
specified decision tables may be defined in a variety of different
ways. The technique was tested with a real database  and the
results with the HIV database are acceptable with accuracies
ranging from 74.7\% to 100\%. One drawback of this method is that
it makes no extrapolation or interpolation and as a result, can
only be used if the missing case is similar or related to another
case with full or more observation.

\section{Acknowledgement}

The financial assistance of the National Research Foundation
(NRF) of South Africa and the Council for Scientific and Industrial Research (CSIR)towards this research is hereby acknowledged.





\bibliographystyle{elsart-harv}
\bibliography{refsTHESIS}

\end{document}